\pdfoutput=1

\documentclass[11pt]{article}

\usepackage{acl}

\usepackage{times}
\usepackage{latexsym}
\usepackage{amsmath}

\usepackage{graphicx}
\usepackage{cleveref}
\usepackage{multirow}
\usepackage{booktabs}
\usepackage{paralist}

\usepackage{tikz}
\newcommand*\circled[1]{\textcircled{{\footnotesize #1}}}
\usepackage{array}

\newcolumntype{H}{>{\setbox0=\hbox\bgroup}c<{\egroup}@{}} 

\usepackage[T1]{fontenc}
\usepackage{todonotes}

\usepackage[utf8]{inputenc}

\usepackage{microtype}

\crefformat{section}{\S#2#1#3}
\crefformat{subsection}{\S#2#1#3}
\crefformat{subsubsection}{\S#2#1#3}
\crefmultiformat{section}{\S#2#1#3}{ and~\S#2#1#3}{, \S#2#1#3}{, and~\S#2#1#3}
\crefmultiformat{subsection}{\S#2#1#3}{ and~\S#2#1#3}{, \S#2#1#3}{, and~\S#2#1#3}
\crefmultiformat{subsubsection}{\S#2#1#3}{ and~\S#2#1#3}{, \S#2#1#3}{, and~\S#2#1#3}
\crefrangeformat{section}{\mbox{\S\S#3#1#4--#5#2#6}}
\crefrangeformat{subsection}{\mbox{\S\S#3#1#4--#5#2#6}}
\crefrangeformat{subsubsection}{\mbox{\S\S#3#1#4--#5#2#6}}

\title{Empirical Sufficiency Lower Bounds for Language Modeling with Locally-Bootstrapped Semantic Structures}

\author{Jakob Prange \and Emmanuele Chersoni\\
Department of Chinese and Bilingual Studies, The Hong Kong Polytechnic University \\
11 Yuk Choi Road, Hung Hom, Kowloon, Hong Kong (China)\\
  {\tt first.last@polyu.edu.hk}}

\begin{document}

\maketitle

\begin{abstract}
    In this work we build upon negative results from an attempt at language modeling with predicted semantic structure, in order to establish empirical lower bounds on what \textit{could} have made the attempt successful.\footnote{Our experimental code is available at \url{https://github.com/jakpra/SufficiencyLowerBounds}.}
    More specifically, we design a concise binary vector representation of semantic structure at the lexical level and evaluate in-depth how good an incremental tagger needs to be in order to achieve better-than-baseline performance with an end-to-end \textit{semantic-bootstrapping language model}. We envision such a system as consisting of a (pretrained) sequential-neural component and a hierarchical-symbolic component working together to generate text with low surprisal and high linguistic interpretability.
    We find that (a) dimensionality of the semantic vector representation \textit{can} be dramatically reduced without losing its main advantages and (b) lower bounds on prediction quality cannot be established via a single score alone, but need to take the \textit{distributions} of signal and noise into account.
\end{abstract}

\section{Introduction}\label{sec:intro}

It is well-established by now that large pretrained Transformer language models (LMs) can obtain detectable knowledge about linguistic structure from raw text distributions \citep[][\textit{inter alia}]{jawahar-etal-2019-bert,tenney-etal-2019-bert}, thus continuing a long line of research in collecting solid empirical evidence for the Distributional Hypothesis \citep{harris_distributional_1954,firth_synopsis_1957}. 
This is often presented in stark contrast to symbolic linguistic theories and representations, which put more emphasis on higher-level structural principles.
In practice, purely neural models have achieved groundbreaking performances in a wide range of NLP tasks \citep{devlin-etal-2019-bert,brown2020language} in a much more scalable manner than seems to be possible with symbolic ones.
Still, theoretical linguistic questions about the relationship between neural implementation and higher-level symbolic patterns are far from being answered definitively.
A common criticism of purely distributional models is that they generally \textit{lack grounding}, because they do not have access to the external world, while \textit{meaning} is inherently a relation between a linguistic form and a communicative intent about something external to language \citep{bender-koller-2020-climbing,trott-etal-2020-construing,merrill-etal-2021-provable,lenci2023understanding}.\footnote{But see \citet{abdou-etal-2021-language} for a more optimistic view, backed by empirical results.}

\begin{figure}
    \centering\small
    \includegraphics[width=\columnwidth]{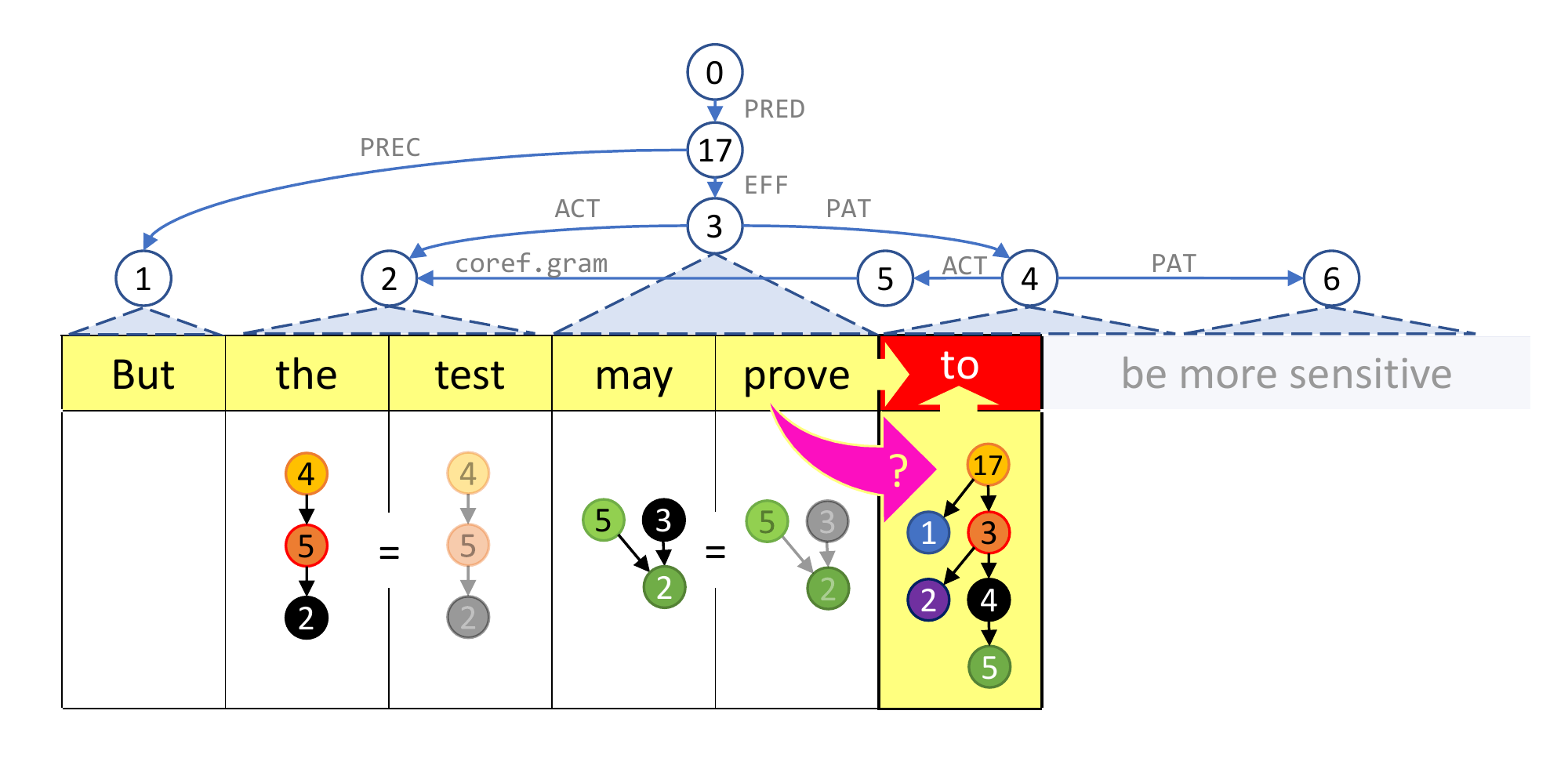}
    \caption{Example of incremental semantic graph slices obtained from a PTG graph and information flow in a (hypothetical) semantic-bootstrapping LM. 
    In this example, the dark-red-shaded token `to' is the current LM target; the light-yellow-shaded cells to the left and below directly influence the LM decision as in P+22; and the pink arrow marked with \textsf{?} stands for the intermediate slice prediction step, for which we want to establish sufficiency lower bounds.}
    \label{fig:slices}
\end{figure}

We aim to contribute to this discussion by building upon results by \citet[henceforth P+22]{prange-etal-2022-linguistic}, who found that incremental LM perplexity can be significantly improved by providing hierarchical semantic structure as an additional token-level input (\cref{fig:slices,sec:slices}).
Indeed, the integration of symbolic and distributional approaches has long been seen as a possible and necessary step towards the full legitimacy of Distributional Semantic Models (DSMs) as models of meaning \citep{boleda2016formal,emerson-2020-goals}, and there is recently more and more evidence supporting the benefits of hybrid neuro-symbolic models \citep[e.g.,][]{li-srikumar-2019-augmenting,li-rush-2020-posterior}, especially for compositional and long-tail generalization \citep{weissenhorn-etal-2022-compositional,prange-etal-2021-supertagging} and interpretability \citep{opitz-frank-2022-sbert}.

P+22's results seem to suggest that at least some aspects of symbolic semantic structure may not be contained in the incremental LM's representation---i.e., that these aspects might constitute an instance of grounding, which is helpful for language understanding, but not fully learnable from text alone.
Alternatively, we consider the possibility that the crucial semantic information \textit{could} be learned, extracted, or induced to a sufficient extent, if only explicit supervision were provided at training time.
The notion of \textit{sufficiency}, in our case, relates to the potential of improving over a baseline LM (\cref{sec:lower-bounds}).
This paints a grand vision of semantic bootstrapping, i.e., a scenario in which the LM first commits to local semantic structure based on the revealed sentence prefix (pink \textsf{?} arrow in \cref{fig:slices}) and then uses its prediction to reduce the next token's perplexity.
The work by P+22 established \textit{upper bounds} by using an oracle setup where rich semantic structure inputs are known to be correct, not only during training but also at test time.
As the main contribution of this work, assuming the local semantic bootstrapping scenario is feasible at all, we look instead for \textit{lower bounds} on what would constitute \textit{sufficient substance and accuracy} in predicted semantic structure for such an improved end-to-end neuro-symbolic LM.

Concretely, we conduct two analyses:
First, we make P+22's original formulation of semantic graph slices (SGS) more parsimonious (\cref{sec:simplify}).
We extract binary vectors (B-SGS) representing only bare-bones (unlabeled and unanchored) structural relations (\cref{sec:simplify-method}) and find that they are sufficient for improving LM perplexity over a strong baseline in the oracle setting (\cref{sec:simplify-eval}). 
Second, we measure how the language modeling benefits of B-SGS are affected by increasing levels of noise, aiming to emulate various imperfect taggers (\cref{sec:noise}).
Interestingly, a comparison of two different shuffling mechanisms (\cref{sec:noise-shuffle}) as well as a simple pilot tagger (\cref{sec:noise-pred}) reveals that \textit{how errors are distributed} throughout the data is much more important than overall error rate.
Based on our observations, we establish sufficiency lower bounds of B-SGS for use in a semantic bootstrapping LM.
We begin by providing the reader with relevant background information from the literature (\cref{sec:background}), defining concisely what we mean by \textit{sufficiency lower bounds} (\cref{sec:lower-bounds}), and describing our data set and general experimental setup (\cref{sec:exp-setup}).
Finally, we discuss our findings and limitations within the bigger picture of ongoing research directions (\cref{sec:discussion}).

\section{Background}\label{sec:background}

\subsection{Language Modeling with Linguistic Graph Slices}\label{sec:slices}

P+22 proposed a type of ensemble language model, consisting of a pretrained Transformer and a neural encoder of symbolic linguistic structure, both jointly predicting the next token in a sentence, given the revealed prefix.
They extract token-level ``slices'' from sentence-level graphs. 

An incremental linguistic graph \textbf{slice} is defined as a connected subgraph minimally including a node directly anchored in the target token (or a preceding token if no such node is available) and extending vertically to include parents, grandparents, and children, horizontally to include left siblings, and diagonally to include children's parents (``co-parents'') and parents' siblings (``aunts''). This is illustrated in \cref{fig:slices}: The original sentence-level graph is shown above the sentence, and extracted token-level slices are shown below.
Slices are then \textbf{encoded} as fixed-length vectors, including both edge label information and token anchor information. Out of two encoding methods, R\mbox{-}GCN \citep{schlichtkrull-etal-2018-rgcn} and a simple concatenation- and averaging-based one, the latter is much faster at roughly equal model size and roughly equal LM quality, so we choose it in our experiments. In essence, the embeddings of all preceding tokens related to the target token in one of the structural ways listed above (parents, siblings, etc), as well as their one-hot-encoded edge labels, are concatenated in a specific pre-defined order. If there are multiple instances of a given relation, or multi-token anchors, their vector representations are averaged. Missing relations are zero-padded.
The final slice vector is fed through a simple feed-forward encoder in order to compute logits over the vocabulary, which are finally added to the LM's logits before softmax normalization. The resulting \textbf{distribution} is used to compute the loss (during training) or predict the next token (at test time).

In their study, P+22 compared linguistic representations of several different flavors, including syntactic and dependency frameworks. Here we focus on two semantic frameworks, PTG and EDS (\cref{sec:data}), which structurally go beyond bilexical dependencies, and thus we use the term \textit{semantic graph slice} (SGS).
We further extend P+22's work by explicitly comparing their oracle setup against several versions of SGS with varying degrees of richness and correctness, stemming from either signal reduction (\cref{sec:simplify}), automatic prediction (\cref{sec:noise-pred}), or controlled noise induction (\cref{sec:noise-shuffle}).

\subsection{Related Work}\label{sec:rel-work}

\paragraph{Linguistic Analyses of LMs.}

A large number of studies in the LM literature has been dedicated to the analysis of the linguistic knowledge they encode. A common methodology employs \textit{probing tasks}, where a simple model is asked to solve a task requiring linguistic knowledge using a representation derived from a LM with little or no specific linguistic supervision. If the model is successful, we then can infer that the LM encodes that knowledge \citep[see][\textit{inter alia}]{linzen-etal-2016-assessing,tenney-etal-2019-bert,tenney2018what,hewitt-liang-2019-designing,liu-etal-2019-linguistic,wu-etal-2020-perturbed,chersoni2021decoding,Geiger:Lu-etal:2021-abstractions}. 
Probes can be particularly insightful when applied \textit{contrastively} to sets of minimal sentence pairs that differ in their grammatical acceptability \citep{warstadt-etal-2020-blimp-benchmark,hu-etal-2020-systematic,kim-etal-2019-probing}.
Our approach of treating semantic structure as an input rather than an output of a neural LM is orthogonal to probing, but can similarly be used for inferences about what kind of knowledge is (not) already encoded in the baseline model.
Recently, an interpretability method based on contrastive \textit{explanations} \citep{jacovi-etal-2021-contrastive} has been proposed to explain LM predictions on sets of minimal sentence pairs that differ in their grammatical acceptability, showing that the salient tokens for the LM preference of the correct form are quite well aligned with human knowledge of grammatical phenomena \citep{yin-neubig-2022-interpreting}.

\paragraph{Incremental Supertagging and Parsing.}

Predicting linguistic structure incrementally has been explored especially in the context of strongly-formulated lexico-syntactic grammar formalisms like CCG, in the form of incremental supertagging \citep{hassan-etal-2009-lexicalized,ambati-etal-2015-incremental,stanojevic-steedman-2019-ccg,stanojevic-steedman-2020-max}.
Having word-level structural categories built in to the formalism has many advantages for both modeling efficiency and linguistic interpretability.
But also Penn Treebank-style constituency syntax trees can be parsed incrementally using, e.g., language model grammars \citep{sartran-etal-2022-transformer,dyer-etal-2016-recurrent} or word-level beam search \citep{stern-etal-2017-effective}.
Finally, another line of work aims to backpropagate linguistic knowledge into the LM itself by optimizing incremental structure prediction as an auxiliary objective \citep{qian-etal-2021-structural,glavas-vulic-2021-supervised,kitaev-etal-2022-learned}.

\paragraph{Model Explanations and Cognitive Predictions using Linguistic Symbols.}

\citet{hale-etal-2018-finding} proposed a method relying on probabilistic generative grammars \citep{dyer-etal-2016-recurrent} and word-synchronous beam search that allows to extract predictive metrics of processing difficulty, such as surprisal and entropy. The authors showed that, using such metrics as predictors in a regression model, it was possible to predict the amplitude effects of several components of naturalistic EEG.
\citet{ek-etal-2019-language} enhance a LSTM-based LM with syntactic, semantic tags and dependency tree depth features, and reported that the additional linguistic knowledge did not increase the correlation with human ratings in a sentence acceptability task, although syntactic tags and dependency tree depth were helpful for lowering perplexity.
\citet{stanojevic-etal-2021-modeling} used CCG-based predictors to improve a regression model of fMRI time course in six different brain regions, over and above predictors obtained with a simple context-free phrase structure grammar.
Finally, \citet{opitz-frank-2022-sbert} presented a technique to partition the BERT sentence embeddings into different sub-embeddings, each one covering meaningful semantic aspects of sentences as represented in the Abstract Meaning Representations (AMR) framework. Experiments on zero-shot sentence and argument similarity tasks proved that the approach maintains a high-level of correlation with human judgements, while making the sentence embeddings interpretable.

\section{Sufficiency Lower Bounds}\label{sec:lower-bounds}

We introduce the concept of \textit{sufficiency\footnote{We do not consider \textit{necessity} lower bounds here. I.e., we do not say that data signals of worse substance than our lower bounds cannot be sufficient. We say that distributions of at least lower-bound quality are probably sufficient.} lower bounds} on the strength of a data signal $\xi$ in order for a system $\Sigma$, which takes $\xi$ as an input, to reach a certain performance threshold $\theta$. 
In this work, the system $\Sigma$ is a neuro-symbolic LM as proposed by P+22 (\cref{sec:slices}), $\xi$ is an SGS vector representation (\cref{sec:simplify}) for each (sub)word token in a text corpus $D$, and $\theta$ is the baseline LM performance (measured as surprisal, \cref{sec:eval}).
Establishing such bounds is important because $\xi$'s richness may need to be reduced in one way or another---either by theoretical design (because small, simple representations and models are desirable; \cref{sec:simplify}), or by practical necessity (due to unavoidable noise in predicting $\xi$; \cref{sec:noise}).
A main takeaway from our exploratory study is that it is important to identify (i.e., define or measure) candidate bounds in a way that considers the signal's configuration as a whole, rather than focusing on a single aggregate metric.\footnote{This somewhat circular-looking reasoning warrants full disclosure: We were already proponents of holistic, detailed evaluations over single-number benchmarks before this study, but were still surprised by most of our results, particularly the contrast between \cref{sec:noise-pred,sec:noise-shuffle}.}
Approached empirically, this involves computing (multivariate) distributions over $\xi$ as instantiated in a data set $D$, such that when the system $\Sigma$ is run on $D$, the quality of its output is at least $\theta$ (i.e., it outperforms a baseline).
Simply put, if the signal $\xi$ surpasses the sufficiency lower bound in $D$, it will likely enable the system $\Sigma$ to reach performance $\theta$ or better.

\section{Experimental Setup}\label{sec:exp-setup}

\subsection{Data}\label{sec:data}

We use the jointly-annotated corpus of the cross-framework meaning representation parsing (MRP) shared tasks \citep{oepen-etal-2019-mrp,oepen-etal-2020-mrp}, which consists of large parts of the English Wall Street Journal corpus.
In particular, we examine two symbolic-structured linguistic representation frameworks, Prague Tectogrammatical Graphs \citep[\textbf{PTG};][]{sgall-etal-1986-meaning,bohmova-etal-2003-prague,hajic-etal-2012-announcing} and Elementary Dependency Structures \citep[\textbf{EDS};][]{oepen-lonning-2006-discriminant,flickinger-2000-building,copestake-etal-2005-minimal}, each of them focusing on different aspects of semantic predicate-argument structure.
EDS derives from Minimal Recursion Semantics (MRS) and thus rather explicitly encodes nominal\slash referring expressions due to MRS' foundation in variable binding.
PTG, on the other hand, is somewhat more guided by syntax and (case\mbox{-})semantic roles.
We use the same training split as P+22, but deviate slightly in using only the first 500 sentences of their development set and reporting most of our results and analyses on this subset. This is because we are reporting incremental results and wish to reserve substantial unseen data for unbiased full evaluation in future work.
For comparison, we report a limited amount of aggregate scores over the test set in \cref{tab:gold-valid}.

\subsection{Model Implementation}

Our models (see \cref{sec:slices} for a conceptual overview) and experiments are implemented in Python, building on P+22's codebase.\footnote{\scriptsize{\url{https://github.com/jakpra/LinguisticStructureLM}}}
In addition to standard neural language modeling libraries used therein (PyTorch, huggingface), we also leverage the Pyro\mbox{-}PPL library \citep{bingham2018pyro} to implement the variational autoencoder (\cref{sec:noise-pred}).

We follow P+22 in using GPT-2 \citep[][124M parameters]{radford2019language} as the pretrained incremental language model and a simple multilayer perceptron (MLP) to encode and project slice vectors into the vocabulary. These logits are then added to the LM's before taking the softmax to obtain the final next-token prediction distribution.
During training, tokens are sampled from a categorical distribution and contribute to the VAE's overall ELBO loss. While this technically is a slight difference to P+22, who used categorical cross-entropy loss, we are able to closely reproduce their reported baseline perplexity on the test set ($\approx$ 46 $\pm$ 0.1).
As the language modeling baseline we finetune GPT-2 in the target domain (on the raw WSJ text) without any exposure to SGS, as did P+22.

\subsection{Evaluation}\label{sec:eval}

We measure language modeling performance in terms of surprisal or \textit{perplexity} (PPL), which is computed as the exponential of the model's token-averaged negative log-likelihood (NLL).\footnote{See Limitations section for shortcomings of this metric.}
Whenever we report aggregate performance over all data, we use PPL (\cref{tab:gold-valid,tab:shuffle}), but in the detailed analysis of smaller subsets of data we switch to NLL for better readability (\cref{fig:perf-distr}).
For both metrics, lower is better.
To evaluate B-SGS correctness, we consider binary micro-accuracy over individual vector dimensions, macro-accuracy over tokens, as well as edge precision, recall, and F1-score.

\section{Representation Distillation: What makes semantic structure valuable to language modeling?}\label{sec:simplify}

Currently well-known as a popular and effective deep learning technique \citep[e.g.,][]{polino2018model,sanh2019distilbert}, distillation (of neural models) aims to reduce redundancy and unwieldiness (\cref{sec:simplify-motivation}) while retaining core information.
Here we apply a similar concept to a family of \textit{symbolic linguistic} representations, SGS.
Rather than relying on a blackbox training process to transfer knowledge from a large pretrained model to a smaller model, we manually design a less detailed variant of SGS, which we call B-SGS (\cref{sec:simplify-method}). We use ground-truth B-SGS as additional input to the incremental LM as before and find that it \textit{does} constitute a \textbf{lower bound of sufficient richness} (\cref{sec:simplify-eval}).

\subsection{Unparsimoneousness of Fully Labeled and Anchored SGS}\label{sec:simplify-motivation}

\begin{figure}[t]
    \centering\small
    \includegraphics[width=\columnwidth]{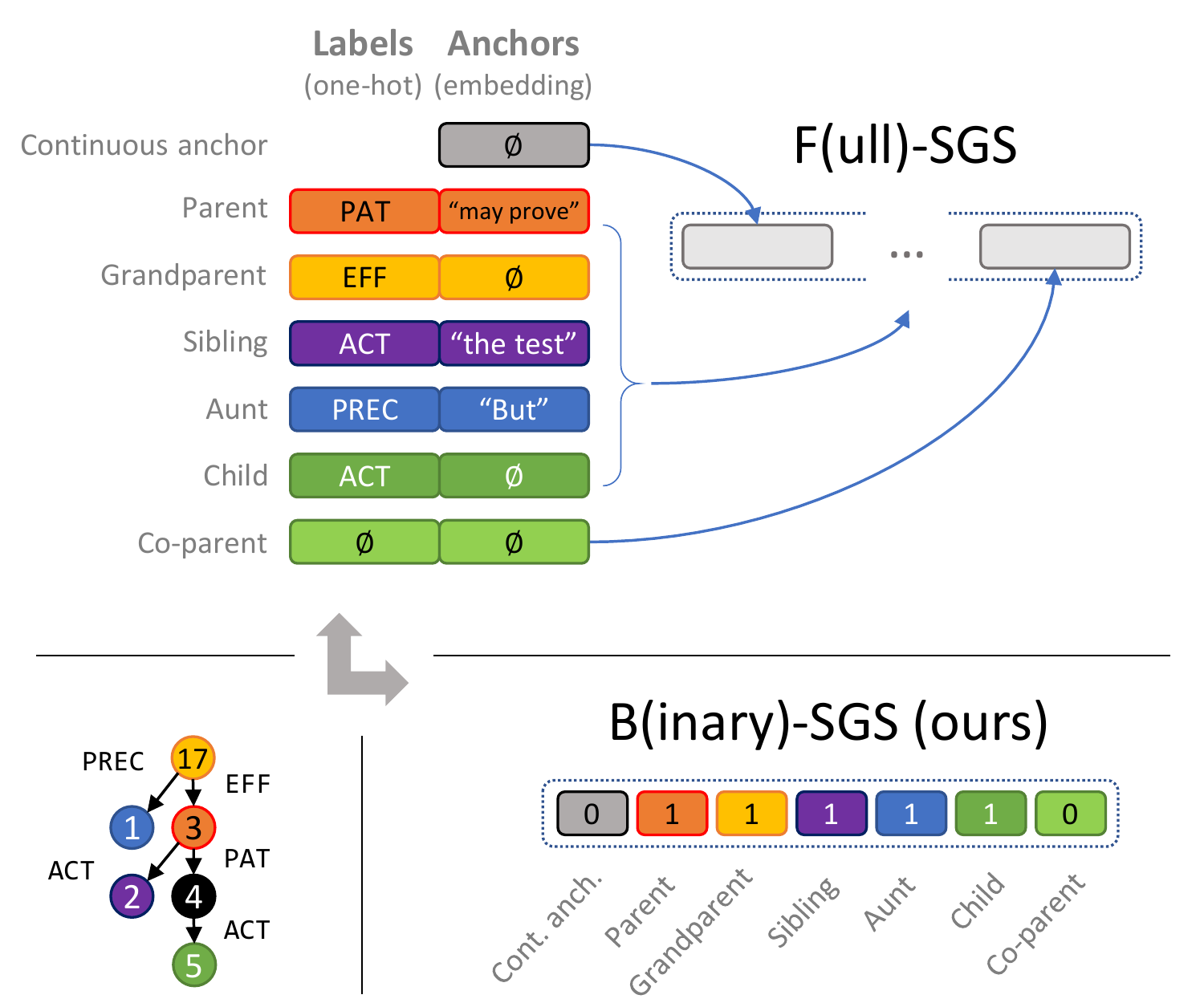}
    \caption{Deriving \textsf{\textbf{F}}ull and \textsf{\textbf{B}}inary semantic graph slice (SGS) vectors from the PTG subgraph for the token `to' in \cref{fig:slices}. The \textit{continuous anchor} dimension would be filled, e.g., in the SGS for the tokens `test' and `prove', which each share the rest of their slice with their respective preceding tokens. Node \circled{5} in the slice for `may prove' is an example of a co-parent.}
    \label{fig:binary_slice}
\end{figure}

While the very rich SGS representation used by P+22 (which, here, we call F(ull)-SGS; \cref{sec:slices} and \cref{fig:binary_slice} top) proved to be a very potent next token predictor, this power comes at the cost of being rather unwieldy and, as it turns out, redundant.

\paragraph{As input.}

Recall from \cref{sec:slices} that, in F-SGS, preceding tokens that are semantically related to each target token (via edges in the graph) are encoded by concatenating their embeddings (in a specific order and with zero-padding to preserve their structural relation, e.g., parent vs. sibling, see \cref{sec:slices}).
It is obvious at first glance that this quickly leads to very large slices and models (P+22 report an average influx in models size of 50-60 million parameters for SGS encoding alone).
Furthermore, linguistic formalisms vary greatly in the number of semantic relation types (edge labels) they distinguish: e.g., 10 in EDS vs. 72 in PTG.
And while this number does not seem to be directly associated with model performance, it still makes the comparison somewhat blurry.
In addition to their excessive size, F-SGS vectors also seem to be partially redundant with a pretrained LM, since P+22 found in their ablation experiments that the correct edge label assignment is not essential for achieving high language modeling performance.

\paragraph{As output.}

In addition to oracle-augmented language modeling, a major use case of SGS we work towards is to incrementally predict them (cf. \cref{sec:intro,sec:noise}).
This is, however, a non-trivial structured prediction problem. It consists at least of edge prediction and relation classification \citep[cf.][]{liu-etal-2019-linguistic}.
And while on the surface, this is reminiscent of a task that could be solved with an edge-factored parser \citep{kiperwasser-goldberg-2016-simple,dozat2017deep}, our scenario is much more complex due to the multitude of structural relations (not just parents), the possibility of multiple parents for each node, abstract nodes not directly anchored in a single token, as well as incrementality. Indeed, it is more akin to supertagging \citep[][\cref{sec:rel-work}]{bangalore-joshi-1999-supertagging,clark-curran-2004-importance} but without the formal guarantees of a mildly context-sensitive grammar formalism like TAG or CCG.
In our early exploration with simple multilayer perceptron (MLP) classifiers and a combination of loss functions (categorical cross-entropy for labels; cosine similarity and\slash or attention loss for token-to-token anchoring), we found it very difficult to train a model to convergence. We suspect that full SGS prediction warrants more complex modeling, optimization, and inference mechanisms, which we leave to future work.

\subsection{Reducing SGS to Binary Structural Relations}
\label{sec:simplify-method}

The challenges described above prompt us to drastically simplify the SGS encoding.
We propose to collapse both edge labels and anchor-token embeddings into mere binary indicators of whether an edge of a given \textit{structural} relation type (flat subword continuation, parent, sibling, grandparent, aunt, child, co-parent) exists, resulting in \textit{binary} semantic graph slices, or B-SGS (\cref{fig:binary_slice}).\footnote{While each node may have multiple relatives of the same type (e.g., 2 parents and 3 siblings), a single binary dimension for each type could only indicate the existence of \textit{at least} one such instance. We follow P+22 in allocating additional `low-resolution' dimensions for certain relation types to indicate the existence of 2 or more relatives. This is illustrated for parents (P+) in \cref{tab:binary-stats} but otherwise omitted (e.g., from \cref{fig:binary_slice}) for simplicity. Note that a node having multiple parents is distinct and independent from it having one or more \textit{co-parents} (i.e., other parents of the node's children).}

\paragraph{Data statistics.}

\begin{table}[t]
    \centering\small
    \begin{tabular}{c cccccccc}
            & A & P & P+ & O & S & T & C & R \\\midrule
        UD & .11 & .34 & 0. & .24 & .31 & .17 & .34 & 0. \\
        DM & .33 & .48 & .21 & .19 & .24 & .09 & .25 & .21 \\\midrule
        PTG & .43 & .75 & .07 & .69 & .42 & .41 & .41 & .10 \\ 
        EDS & .30 & .69 & .51 & .17 & .29 & .09 & .26 & .22 \\
        \bottomrule
    \end{tabular}
    \caption{Relation-wise density of B-SGS vectors in the development set. A: continued anchor, P: parent, P+: 2 or more parents, O: grandparent, S: sibling, T: aunt, C: child, R: co-parent.
    UD and DM are shown for reference \citep[cf.][\cref{sec:noise-pred}]{liu-etal-2019-linguistic}.}
    \label{tab:binary-stats}
\end{table}

We report average density of major SGS dimensions (= relation types) in \cref{tab:binary-stats}. Note in particular that EDS and PTG differ substantially in the types of structures they encode, with PTG being denser on average. EDS is quite similar to DM because they are both derived from the same underlying formalism. In contrast to EDS, PTG, and DM graphs, which are generic DAGs, UD graphs are strictly bilexical dependency trees, leading to necessarily empty P+ and R dimensions.

\subsection{Validating LM Performance with Oracle B-SGS}\label{sec:simplify-eval}

\paragraph{Setup.}

We train for up to 10 epochs, with early stopping based on development set perplexity. See \cref{sec:exp-setup} for more details.

\paragraph{Results.}

\Cref{tab:gold-valid} shows that although B-SGS perplexity is slightly worse than with F-SGS---which is to be expected given the drastic reduction of the input signal---it still clearly outperforms the non-symbolic baseline.
This suggests that the most crucial signal contributed by SGS in general is, in fact, the bare-bones hierarchical structure itself. And while P+22's ablation analysis already suggested that the grouping into different edge labels may be less important, it is quite surprising that even the information about \textit{which other tokens} the target token is hierarchically-related to is not necessary to improve language modeling with SGS.

A possible explanation can be found in the fact that the baseline LM already has extensive \textit{gradient} representation of parts-of-speech, syntactic functions, and semantic roles (namely, in its dense hidden states and attention distributions). What it might be lacking, then, is any \textit{discrete} representation, and in particular a commitment to discrete and complex semantic structure seems to be beneficial.

Gold B-SGS is thus a sufficiency lower bound.

\begin{table}[t]
    \centering\small
    \begin{tabular}{l cHH cHH}
  & PTG & \multicolumn{1}{H}{Acc} & \multicolumn{1}{H}{F}           & EDS & \multicolumn{1}{H}{Acc} & \multicolumn{1}{H}{F} \\\midrule

Pretrained GPT-2 & \multicolumn{4}{c}{59.3} \\ 
Domain-finetuned (baseline) & \multicolumn{4}{c}{46.1} \\\midrule

Gold F-SGS & 26.8 & 1.00 & 1.00 & 24.7 & 1.00 & 1.00 \\[3pt]
\textbf{Gold B-SGS} (ours) & 33.9 & 1.00 & 1.00 & 28.0 & 1.00 & 1.00 \\

\bottomrule

    \end{tabular}
    \caption{Comparing test set LM perplexity (lower is better) with our \textbf{B}inary slices against \textbf{F}ully labeled\slash anchored ones (P+22).}
    \label{tab:gold-valid}
\end{table}

\section{Noise Robustness: How accurate should bootstrapped semantic structure be in order to improve a LM?}\label{sec:noise}

In a pilot experiment, we integrate into the P+22 model B-SGS \textit{prediction}. As illustrated in \cref{fig:vae}, this is an intermediate step, the output of which is now used as input to next-token prediction instead of the ground truth slice. 
We find that while our relatively simple model (\cref{sec:noise-pred}) produces B-SGS outputs of seemingly reasonable overall quality (in terms of micro-accuracy and F-score), they are \textit{not} sufficient for supporting LM performance.
This prompts us to actively search for \textbf{lower bounds of sufficient correctness} by artificially inducing various types and levels of noise into gold B\mbox{-}SGS inputs (\cref{sec:noise-shuffle}).
We do find several bounds, but learn that what makes them sufficient has less to do with their single-number correctness and more with intricate details of their overall noise distribution (\cref{sec:perf_analysis}).

\subsection{Pilot Prediction}\label{sec:noise-pred}

\paragraph{Setup.}

\begin{figure}[t]
    \centering\small
    \includegraphics[width=0.8\columnwidth]{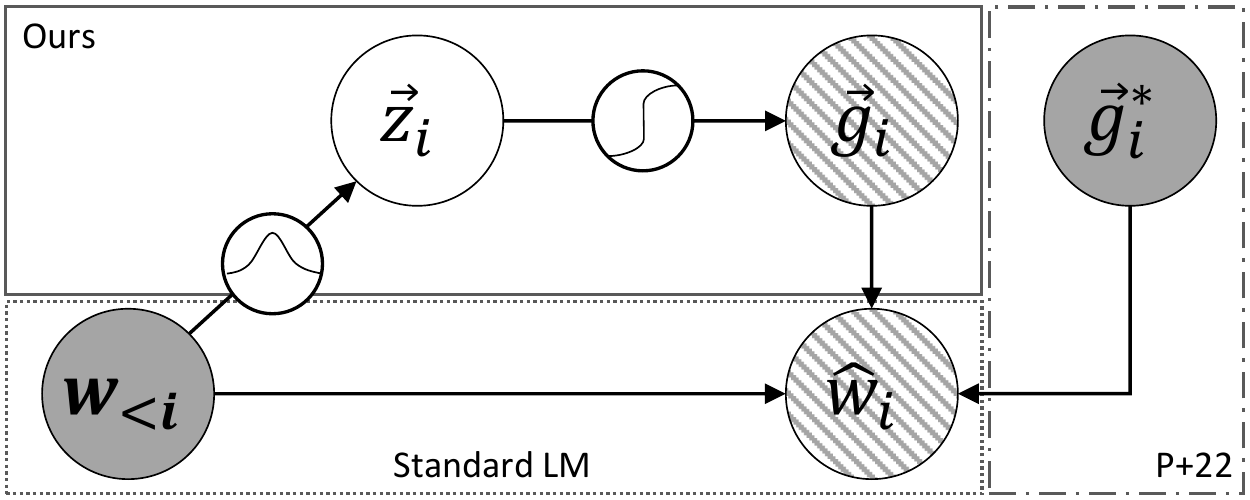}
    \caption{Our simple variational autoencoder model. We project the encoding of observed (solidly shaded) previous words $\bf{w}_{<i}$ into latent space and sample hidden states $\vec{z}_i$. Predicted graph slices ${\vec{g}}_i$ and target tokens $\hat{w}_i$ are supervised during training but unobserved at test time. P+22 used ground truth slices $\vec{g}^*_i$ instead of predicted ones. The standard LM is a component in both versions.}
    \label{fig:vae}
\end{figure}

Since we are interested in lower bounds and we are running an exploratory study, we do not perform extensive model engineering. The following description is purely intended for clarity and replicability rather than as a state-of-the-art model proposal. 
We decide on a variational autoencoder \citep[VAE;][]{kingma2013auto}, where sampling from the latent space mediates between the LM's hidden state and the sigmoid-activated B-SGS dimensions (\cref{fig:vae}). This setup is motivated by the high uncertainty involved in the task (we predict the symbolic structure of a token that has not been observed yet, and there may be much genuine ambiguity). All encoders, decoders, and projectors within the VAE, besides GPT-2, are simple feed-forward MLPs.
B-SGS prediction is trained deterministically with binary cross-entropy loss.\footnote{We also experimented with Bernoulli sampling, but to no success.}
We train the slice predictor for up to 10 epochs with early stopping based on dev set F-score, and then train the SGS-augmented LM as before.

\paragraph{Results.}

\begin{figure}[t]
    \centering\small
    \includegraphics[width=\columnwidth]{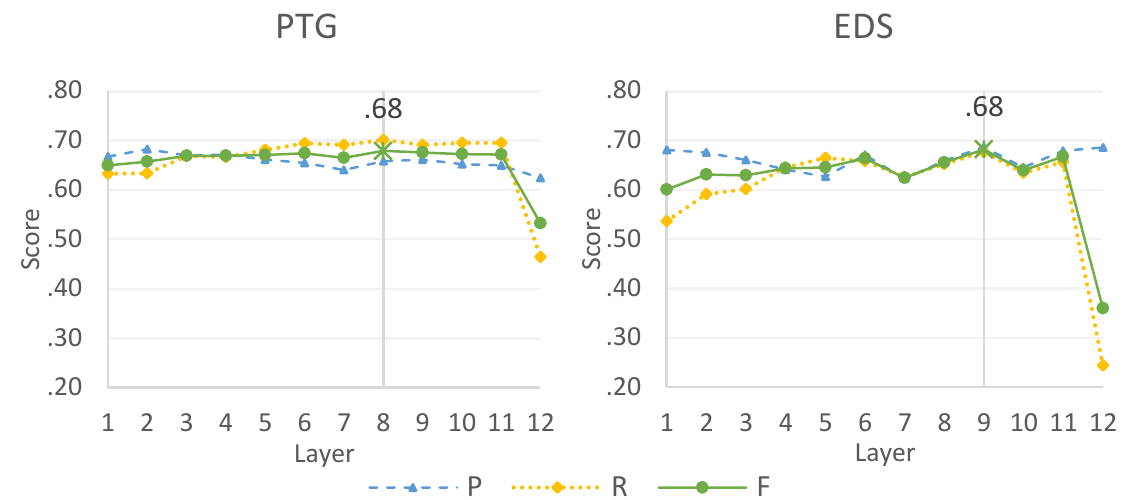}
    \caption{Graph slice prediction performance on the development set by LM layer.
    P = precision (proportion predicted edges correct), R = recall (gold edges predicted), F = F1-score (harmonic mean of P and R).}
    \label{fig:layer-perf}
\end{figure}

As shown in \cref{fig:layer-perf}, SGS prediction performance is best in layers 8 (PTG) and 9 (EDS). This is in line with previous studies on probing semantic structure \citep[e.g.,][]{liu-etal-2019-linguistic,jawahar-etal-2019-bert,tenney-etal-2019-bert}, which obtained the best performances in middle\slash high layers.
However, even these best predictions cannot outperform the finetuned LM baseline in the augmented language modeling setting (compare black solid and red dashed lines in \cref{fig:shuffle-ppl}).

\paragraph{Validation.}

Prediction micro-accuracies (.84 for PTG, .90 for EDS; last row \cref{tab:shuffle}) are in the same order of magnitude as \citet{liu-etal-2019-linguistic}'s binary edge prediction results for UD and DM, two representation frameworks featured in the literature much more frequently than PTG and EDS. Although there are many differences in task and experimental setup (dependencies vs. constituencies, single-parent vs. B-SGS prediction; cf. \cref{tab:binary-stats}), we find this to be a valuable sanity check for both us and the reader in lieu of a proper baseline.

\subsection{Artificial Noise}\label{sec:noise-shuffle}

Why is our pilot system not sufficient? Maybe prediction accuracy just needs to be better?
We investigate by using shuffled gold B-SGS as inputs to the LM and systematically altering several characteristics of the shuffling routine.
This style of control task is inspired by \citet{hewitt-liang-2019-designing,dubossarsky-etal-2018-coming}.

We consider two different shuffling mechanisms: 
(a) Shuffling the node-to-word \textbf{anchor} mapping of graphs before vector extraction (i.e., which slice corresponds to which word token, cf.~P+22).
This guarantees well-formed graphs but may be too optimistic since we only shuffle within each sentence. Thus we also consider a more aggressive option:
(b) Randomly switching \textbf{bits} (= whether or not a given edge type exists) in the slice vectors extracted from gold graphs.

For both, we also produce \textit{varying degrees} of noise.
Namely, whenever we are about to shuffle a graph anchor or vector bit, we decide to instead retain the correct assignment with probability $p_{Gold}$.

\paragraph{Results.}

\Cref{tab:shuffle} shows how the different shuffling conditions affect B-SGS correctness.
As expected, within-sentence graph anchor shuffling is generally much more optimistic than bit-switching. By definition, $p_{Gold}$ directly determines micro-accuracy in bit-switched slices, whereas in anchor-shuffled slices, $p_{Gold}$ is more closely correlated with macro-accuracy.
LM perplexity of each condition is shown in \cref{fig:shuffle-ppl}. Note that the signal strength of bit-switching is symmetric around .5. This is an intuitive corollary of it being a binary signal (though macro-accuracy and F-score naturally continue to decline with $p_{Gold} < .5$, as shown exemplarily for values .1 and 0.).

First, we identify conditions that beat the domain-finetuned LM baseline from \cref{fig:shuffle-ppl}, and then consult \cref{tab:shuffle} to find their corresponding slice quality.
This results in the following sufficiency lower bounds (marked with asterisks in \cref{tab:shuffle}): \textbf{Shuffled graphs} with $p_{Gold} \in \{.9, .8, .7, .6, .5\}$ for both PTG and EDS as well as $p_{Gold} \in \{.4, .3, .2, .1, 0.\}$ for EDS; and \textbf{bit-switched vectors} with $p_{Gold} \in \{.9, .8, .2, .1, 0.\}$, which can be generalized $|p_{Gold} - .5| \geq .3$.

\subsection{Detailed Analysis}\label{sec:perf_analysis}

\begin{figure}[t]
    \centering
    \includegraphics[width=\columnwidth]{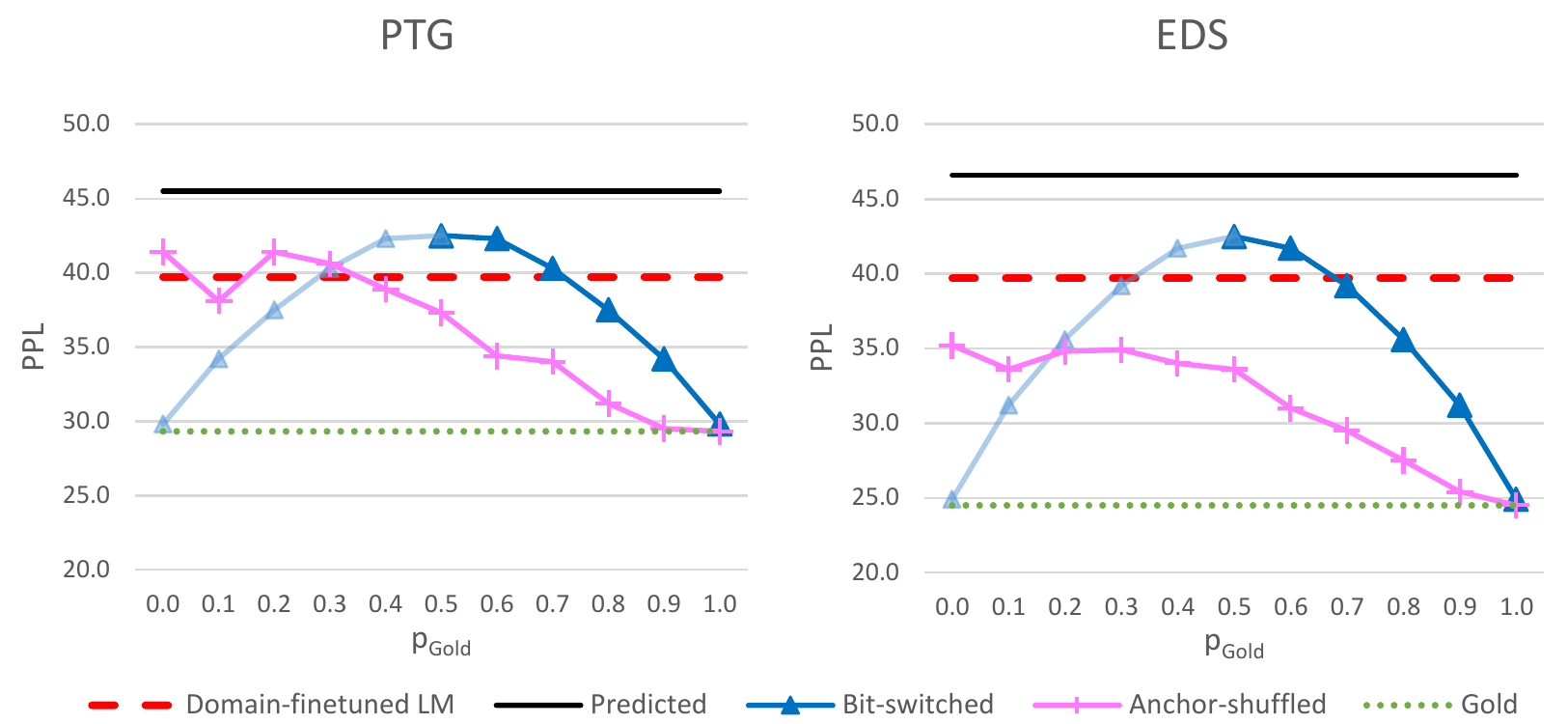}
    \caption{Dev set perplexity (lower is better) of noise-interpolated B-SGS LMs.}
    \label{fig:shuffle-ppl}
\end{figure}

\begin{table}[t]
    \setlength{\tabcolsep}{2.3pt}
    \centering\small
    \begin{tabular}{r rrrc rrrc}
         & \multicolumn{4}{c}{PTG} & \multicolumn{4}{c}{EDS} \\\cmidrule(r){2-5}\cmidrule(l){6-9}
  & \multicolumn{1}{c}{MaA} & \multicolumn{1}{c}{MiA} & \multicolumn{1}{c}{F}           & LB & \multicolumn{1}{c}{MaA} & \multicolumn{1}{c}{MiA} & \multicolumn{1}{c}{F} & LB \\\midrule

\textbf{Gold} & 1.00 & 1.00 & 1.00 & * & 1.00 & 1.00 & 1.00 & * \\\midrule

\multicolumn{7}{l}{\textbf{Shuffled graph anchors}} \\
$p_{Gold}$ = \phantom{0}.9 & $>$.99 & $>$.99 & $>$.99 & * & .95 & .99 & .97 & * \\
$p_{Gold}$ = \phantom{0}.8 & .88 & .97 & .94 & * & .81 & .97 & .91 & * \\
$p_{Gold}$ = \phantom{0}.7 & .60 & .93 & .86 & * & .65 & .95 & .84 & * \\
$p_{Gold}$ = \phantom{0}.6 & .54 & .92 & .83 & * & .53 & .93 & .76 & * \\
$p_{Gold}$ = \phantom{0}.5 & .36 & .87 & .72 & * & .34 & .88 & .63 & * \\
$p_{Gold}$ = \phantom{0}.4 & .23 & .83 & .64 & ? & .25 & .87 & .56 & * \\
$p_{Gold}$ = \phantom{0}.3 & .18 & .81 & .60 & $-$ & .19 & .85 & .49 & *  \\
$p_{Gold}$ = \phantom{0}.2 & .12 & .80 & .56 & $-$ & .17 & .84 & .48 & * \\
$p_{Gold}$ = \phantom{0}.1 & .10 & .79 & .55 & ? & .14 & .83 & .44 & * \\
$p_{Gold}$ = 0. & .08 & .78 & .53 & $-$ & .13 & .82 & .41 & * \\\midrule

\multicolumn{7}{l}{\textbf{Bit-switched vectors}} \\
$p_{Gold}$ = \phantom{0}.9 & .17 & .90 & .81 & * & .17 & .90 & .75 & * \\
$p_{Gold}$ = \phantom{0}.8 & .02 & .80 & .66 & * & .02 & .80 & .58 & * \\
$p_{Gold}$ = \phantom{0}.7 & $<$.01 & .70 & .53 & ? & $<$.01 & .70 & .44 & ? \\
$p_{Gold}$ = \phantom{0}.6 & $<$.01 & .60 & .42 & $-$ & $<$.01 & .60 & .34 & $-$ \\
$p_{Gold}$ = \phantom{0}.5 & 0. & .50 & .32 & $-$ & 0. & .50 & .26 & $-$ \\  
\vdots & & & & \vdots & & & & \vdots \\
$p_{Gold}$ = \phantom{0}.1 & 0. & .10 & .05 & * & 0. & .11 & .05 & * \\
$p_{Gold}$ = 0. & 0. & $<$.01 & $<$.01 & * & 0. & .01 & .02 & * \\
\midrule

\textbf{Predicted} & .18 & .84 & .68 & $-$ & .29 & .90 & .68 & $-$ \\\bottomrule

    \end{tabular}
    \caption{Correctness F1-score (F), accuracy at the macro (token-level, MaA) and micro (bit-level, MiA) levels) of B-SGS with various levels of noise (measured on the dev set). The LB columns indicate whether a condition is a sufficiency lower bound (*=yes, ?=maybe, $-$=no), i.e., if its corresponding PPL beats the baseline (\cref{fig:shuffle-ppl}).}
    \label{tab:shuffle}
\end{table}

\begin{figure*}[t]
    \centering
    \includegraphics[width=\textwidth]{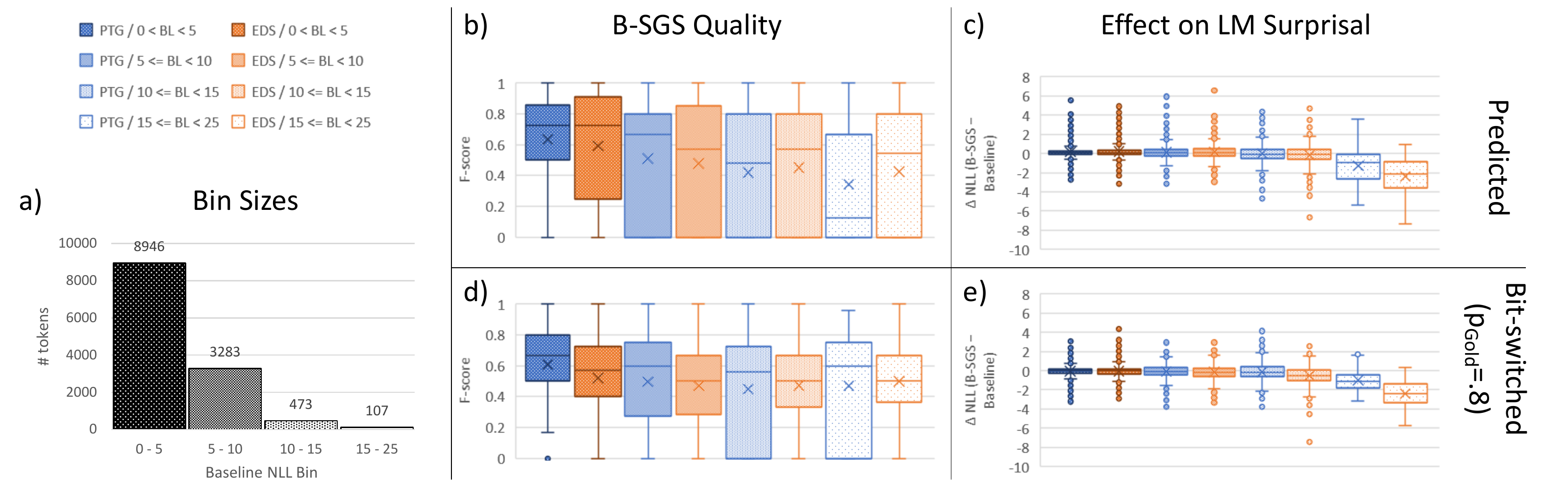}
    \caption{Box-and-whiskers plots of distributions over dev set tokens of B-SGS quality (F1-score) and effect on LM surprisal ($\Delta$ NLL), binned by baseline (BL) NLL. $\times$ markers are means and lines within boxes are medians.}
    \label{fig:perf-distr}
\end{figure*}

Unexpectedly, shuffled slices with clearly worse overall accuracy than predicted ones (\cref{tab:shuffle}) sill yield much better perplexity (\cref{fig:shuffle-ppl}).
This leads us to the following hypotheses which we address in order. For brevity, we focus only the comparison between \textit{predicted} and \textit{bit-switched with $p_{Gold}=.8$}, because this condition seems to be a good trade-off between matching or slightly beating the PPL baseline and realism in terms of closeness to predicted in terms of overall accuracy. Consider \cref{fig:perf-distr}.

\textit{Hypothesis: The noise of shuffled slices is uniformly distributed over tokens whereas the noise of predicted slices is distributed similarly as baseline LM surprisal.}\\
Average F-score of predicted B-SGS does indeed decrease as baseline LM surprisal increases (\cref{fig:perf-distr}b).
However, contrary to our expectation, the \textit{same} is true for the F-score of uniformly shuffled slices (\cref{fig:perf-distr}d)!
Thus, the distribution of F-score means over suprisal bins alone does not explain the difference.

\textit{Hypothesis: Due to high-surprisal tokens having low B-SGS correctness, we create a noisy feedback loop which worsens LM surprisal in particular for already high-surprisal words (open-class content words) without gaining enough advantage on low-surprisal words.}\\
We find quite the opposite: Both predicted and shuffled slices help in particular for very-high-surprisal tokens, despite the higher average slice noise (\cref{fig:perf-distr}c+e). In contrast, predicted slices tend to slightly increase surprisal for low-surprisal tokens.
And since low-BL-surprisal tokens make up the vast majority of the data (\cref{fig:perf-distr}a), this slight increase might be enough to confuse the LM beyond baseline.
Another crucial factor might be \textit{variance} in slice correctness, which is generally much higher in predicted slices than in shuffled ones (\cref{fig:perf-distr}b+d).

\paragraph{Most affected words.}

We manually inspect the data to get an idea of how predicted B-SGS benefits the LM the most.
The top 10 tokens in terms of both baseline NLL and $\Delta$ NLL (bottom right region of \cref{fig:perf-distr}c) are dominated by (recurring) named entities and dates, which are likely just an artifact of overfitting.
After filtering these out, we find that the highest-baseline-NLL tokens are mostly nouns, adjectives, and verbs that are either rare themselves (e.g., \textit{hopscotched}, \textit{instrumentation}) or used in a rare construction (\textit{paying thin \textbf{compliments}}).
In contrast, both PTG and EDS B-SGS reduce NLL the most for verbs, particularly in participle constructions (\textit{dividing}, \textit{has \textbf{begged}}, \textit{will be \textbf{relocated}}).

\section{Discussion and Conclusions}\label{sec:discussion}

We proposed a general framework for semantically-enriched language modeling. Our proposal aims to provide a new perspective on qualitative distributional linguistic analysis, expanding upon prior work in linguistic analysis of neural models in several ways (\cref{sec:framework}).
We implemented and tested this framework with GPT-2 and semantic graph slices (SGS) from two formalisms, finding interesting patterns with potential impact for meaning representation design and low-resource modeling (\cref{sec:concrete}).

\subsection{General Framework}\label{sec:framework}

Probing and related methods evaluate language models based on their ability to predict linguistic representations from text.
Although this is a relatively practical setup which has already produced many fascinating and replicable findings, it has the disadvantage that results need to be interpreted relative to both the linguistic framework governing the output and the specific probing architecture.
In contrast, the approach of \citet{prange-etal-2022-linguistic} takes linguistic representations as an input and evaluates the language model directly on its native language modeling task.
The main problem with their oracle setup is that it is unrealistic to have ground truth linguistic structures available at test time.

We argue for unifying the advantages of both directions, by considering what is in essence a concatenation of the two: a pipeline in which the output of a structure prediction model (similar to probing, except that it may be supervised) is fed back into the LM, enabling comparable evaluation on the raw text itself.
This makes it possible to identify shortcomings of the LM and\slash or benefits of the linguistic representation quantitatively and qualitatively, by modifying either the probing architecture or the linguistic representation itself until LM performance starts or stops improving.
The \textit{lower bounds} of this continuum in particular (in contrast to upper bounds) have many theoretical and practical implications, since they separate the wheat from the chaff when it comes to the efficiency\slash effectiveness trade-off for model and representation.
Our definition of sufficiency lower bounds in terms of the signal's data distribution in \cref{sec:lower-bounds} is intentionally kept high-level and flexible to stimulate adaptations of the idea for a variety of use cases.
While here we take an exclusively empirical approach, the framework may lend itself to formally-provable accounts as well.

\subsection{Concrete Take-aways}\label{sec:concrete}

In our experiments with GPT-2 (\cref{sec:simplify,sec:noise}), we were able to crystallize the simple (unlabeled and unanchored) discrete hierarchical semantic structure of PTG and EDS as both beneficial to language modeling and robust to certain types of noise.
We also found, though, that measuring prediction quality via a single aggregate score hides important aspects of the distributions of signal and noise, to the extent of potentially nullifying LM improvements.
While the respective structures of PTG and EDS differ from each other in terms of density, relations encoded (\cref{sec:simplify-method}), prediction accuracy (\cref{sec:noise-pred}), and LM benefit (\cref{sec:noise-shuffle}), the types of words they help the LM with the most are similar (\cref{sec:perf_analysis}).

As a nice side-effect from \cref{sec:simplify}, removing the explicit token anchoring from SGS also makes it more applicable to unanchored semantic representations such as AMR \citep{banarescu-etal-2013-abstract}. Note, however, that we still need some source of basic anchoring information (e.g., from an automatic aligner) in order to assign a slice to each token.

Finally, based on our findings in \cref{sec:perf_analysis} that rare high-surprisal words most positively affected by even noisily SGS-enhanced language modeling, we are hopeful that our method may be particularly helpful for the Zipfian tail at a small cost to the majority of data.

\section*{Limitations}\label{sec:limitations}

\paragraph{No guarantees.}

As stated in \cref{sec:lower-bounds} and substantiated in \cref{sec:noise}, sufficiency lower bounds tend to be non-trivial, multifaceted configurations. We explore this to some extent (we find, e.g., that overall correctness scores alone, without variance, are \textit{not} reliable identifiers of sufficiency lower bounds), but not exhaustively. To make stronger guarantees rather than just optimism, we need to precisely define when a candidate distribution is `similar enough' to a known lower bound (e.g., via goodness-of-fit).

\paragraph{Practicability of semantic bootstrapping.}

We do not present a complete working system yet.
It could be that sufficiently distributed performance can only be achieved with more intricate structured decoding mechanisms (e.g., Viterbi or beam search), which would negatively affect running time and thus usability as an end-to-end LM.

\paragraph{Limited evaluation of LM quality.} Our evaluation of LM quality has been limited to the effects of the predicted graph slices on the perplexity metric. Alternative evaluations adopting psycholinguistically-inspired metrics, such as the correlation with human norms collected from cloze completion tasks, might yield different results \citep{hao2020probabilistic}.

\section*{Acknowledgements}

We thank the anonymous reviewers for their extremely insightful and constructive feedback and questions, as well as Nathan Schneider and Chu\mbox{-}Ren Huang for helpful early discussions.
This work has been supported by Hong Kong PolyU grant 1\mbox{-}YWBW.

\bibliography{anthology,custom}

\begin{thebibliography}{65}
\expandafter\ifx\csname natexlab\endcsname\relax\def\natexlab#1{#1}\fi

\bibitem[{Abdou et~al.(2021)Abdou, Kulmizev, Hershcovich, Frank, Pavlick, and
  S{\o}gaard}]{abdou-etal-2021-language}
Mostafa Abdou, Artur Kulmizev, Daniel Hershcovich, Stella Frank, Ellie Pavlick,
  and Anders S{\o}gaard. 2021.
\newblock \href {https://doi.org/10.18653/v1/2021.conll-1.9} {Can language
  models encode perceptual structure without grounding? a case study in color}.
\newblock In \emph{Proceedings of the 25th Conference on Computational Natural
  Language Learning}, pages 109--132, Online. Association for Computational
  Linguistics.

\bibitem[{Ambati et~al.(2015)Ambati, Deoskar, Johnson, and
  Steedman}]{ambati-etal-2015-incremental}
Bharat~Ram Ambati, Tejaswini Deoskar, Mark Johnson, and Mark Steedman. 2015.
\newblock \href {https://doi.org/10.3115/v1/N15-1006} {An incremental algorithm
  for transition-based {CCG} parsing}.
\newblock In \emph{Proceedings of the 2015 Conference of the North {A}merican
  Chapter of the Association for Computational Linguistics: Human Language
  Technologies}, pages 53--63, Denver, Colorado. Association for Computational
  Linguistics.

\bibitem[{Banarescu et~al.(2013)Banarescu, Bonial, Cai, Georgescu, Griffitt,
  Hermjakob, Knight, Koehn, Palmer, and
  Schneider}]{banarescu-etal-2013-abstract}
Laura Banarescu, Claire Bonial, Shu Cai, Madalina Georgescu, Kira Griffitt, Ulf
  Hermjakob, Kevin Knight, Philipp Koehn, Martha Palmer, and Nathan Schneider.
  2013.
\newblock \href {https://aclanthology.org/W13-2322} {{A}bstract {M}eaning
  {R}epresentation for sembanking}.
\newblock In \emph{Proceedings of the 7th Linguistic Annotation Workshop and
  Interoperability with Discourse}, pages 178--186, Sofia, Bulgaria.
  Association for Computational Linguistics.

\bibitem[{Bangalore and Joshi(1999)}]{bangalore-joshi-1999-supertagging}
Srinivas Bangalore and Aravind~K. Joshi. 1999.
\newblock \href {https://aclanthology.org/J99-2004} {{S}upertagging: An
  approach to almost parsing}.
\newblock \emph{Computational Linguistics}, 25(2):237--265.

\bibitem[{Bender and Koller(2020)}]{bender-koller-2020-climbing}
Emily~M. Bender and Alexander Koller. 2020.
\newblock \href {https://doi.org/10.18653/v1/2020.acl-main.463} {Climbing
  towards {NLU}: {On} meaning, form, and understanding in the age of data}.
\newblock In \emph{Proceedings of the 58th Annual Meeting of the Association
  for Computational Linguistics}, pages 5185--5198, Online. Association for
  Computational Linguistics.

\bibitem[{Bingham et~al.(2018)Bingham, Chen, Jankowiak, Obermeyer, Pradhan,
  Karaletsos, Singh, Szerlip, Horsfall, and Goodman}]{bingham2018pyro}
Eli Bingham, Jonathan~P. Chen, Martin Jankowiak, Fritz Obermeyer, Neeraj
  Pradhan, Theofanis Karaletsos, Rohit Singh, Paul Szerlip, Paul Horsfall, and
  Noah~D. Goodman. 2018.
\newblock {Pyro: Deep Universal Probabilistic Programming}.
\newblock \emph{Journal of Machine Learning Research}.

\bibitem[{B\"{o}hmov\'{a} et~al.(2003)B\"{o}hmov\'{a}, Haji\v{c},
  Haji\v{c}ov\'{a}, and Hladk\'{a}}]{bohmova-etal-2003-prague}
Alena B\"{o}hmov\'{a}, Jan Haji\v{c}, Eva Haji\v{c}ov\'{a}, and Barbora
  Hladk\'{a}. 2003.
\newblock \href {https://doi.org/10.1007/978-94-010-0201-1_7} {The {P}rague
  {D}ependency {T}reebank: {A} three-level annotation scenario}.
\newblock In Anne Abeill\'{e}, editor, \emph{Treebanks: Building and Using
  Parsed Corpora}, Text, Speech and Language Technology, pages 103--127.
  Springer Netherlands, Dordrecht.

\bibitem[{Boleda and Herbelot(2016)}]{boleda2016formal}
Gemma Boleda and Aur{\'e}lie Herbelot. 2016.
\newblock Formal distributional semantics: Introduction to the special issue.
\newblock \emph{Computational Linguistics}, 42(4):619--635.

\bibitem[{Brown et~al.(2020)Brown, Mann, Ryder, Subbiah, Kaplan, Dhariwal,
  Neelakantan, Shyam, Sastry, Askell et~al.}]{brown2020language}
Tom Brown, Benjamin Mann, Nick Ryder, Melanie Subbiah, Jared~D Kaplan, Prafulla
  Dhariwal, Arvind Neelakantan, Pranav Shyam, Girish Sastry, Amanda Askell,
  et~al. 2020.
\newblock Language models are few-shot learners.
\newblock \emph{Advances in neural information processing systems},
  33:1877--1901.

\bibitem[{Chersoni et~al.(2021)Chersoni, Santus, Huang, and
  Lenci}]{chersoni2021decoding}
Emmanuele Chersoni, Enrico Santus, Chu-Ren Huang, and Alessandro Lenci. 2021.
\newblock {Decoding Word Embeddings with Brain-based Semantic Features}.
\newblock \emph{Computational Linguistics}, 47(3):663--698.

\bibitem[{Clark and Curran(2004)}]{clark-curran-2004-importance}
Stephen Clark and James~R. Curran. 2004.
\newblock \href {https://aclanthology.org/C04-1041} {The importance of
  supertagging for wide-coverage {CCG} parsing}.
\newblock In \emph{{COLING} 2004: Proceedings of the 20th International
  Conference on Computational Linguistics}, pages 282--288, Geneva,
  Switzerland. COLING.

\bibitem[{Copestake et~al.(2005)Copestake, Flickinger, Pollard, and
  Sag}]{copestake-etal-2005-minimal}
Ann Copestake, Dan Flickinger, Carl Pollard, and Ivan~A Sag. 2005.
\newblock Minimal recursion semantics: An introduction.
\newblock \emph{Research on language and computation}, 3(2):281--332.

\bibitem[{Devlin et~al.(2019)Devlin, Chang, Lee, and
  Toutanova}]{devlin-etal-2019-bert}
Jacob Devlin, Ming-Wei Chang, Kenton Lee, and Kristina Toutanova. 2019.
\newblock \href {https://doi.org/10.18653/v1/N19-1423} {{BERT}: Pre-training of
  deep bidirectional transformers for language understanding}.
\newblock In \emph{Proceedings of the 2019 Conference of the North {A}merican
  Chapter of the Association for Computational Linguistics: Human Language
  Technologies, Volume 1 (Long and Short Papers)}, pages 4171--4186,
  Minneapolis, Minnesota. Association for Computational Linguistics.

\bibitem[{Dozat and Manning(2017)}]{dozat2017deep}
Timothy Dozat and Christopher~D. Manning. 2017.
\newblock \href {https://openreview.net/forum?id=Hk95PK9le} {Deep biaffine
  attention for neural dependency parsing}.
\newblock In \emph{International Conference on Learning Representations}.

\bibitem[{Dubossarsky et~al.(2018)Dubossarsky, Grossman, and
  Weinshall}]{dubossarsky-etal-2018-coming}
Haim Dubossarsky, Eitan Grossman, and Daphna Weinshall. 2018.
\newblock \href {https://doi.org/10.18653/v1/D18-1200} {Coming to your senses:
  on controls and evaluation sets in polysemy research}.
\newblock In \emph{Proceedings of the 2018 Conference on Empirical Methods in
  Natural Language Processing}, pages 1732--1740, Brussels, Belgium.
  Association for Computational Linguistics.

\bibitem[{Dyer et~al.(2016)Dyer, Kuncoro, Ballesteros, and
  Smith}]{dyer-etal-2016-recurrent}
Chris Dyer, Adhiguna Kuncoro, Miguel Ballesteros, and Noah~A. Smith. 2016.
\newblock \href {https://doi.org/10.18653/v1/N16-1024} {Recurrent neural
  network grammars}.
\newblock In \emph{Proceedings of the 2016 Conference of the North {A}merican
  Chapter of the Association for Computational Linguistics: Human Language
  Technologies}, pages 199--209, San Diego, California. Association for
  Computational Linguistics.

\bibitem[{Ek et~al.(2019)Ek, Bernardy, and Lappin}]{ek-etal-2019-language}
Adam Ek, Jean-Philippe Bernardy, and Shalom Lappin. 2019.
\newblock \href {https://aclanthology.org/W19-6108} {Language modeling with
  syntactic and semantic representation for sentence acceptability
  predictions}.
\newblock In \emph{Proceedings of the 22nd Nordic Conference on Computational
  Linguistics}, pages 76--85, Turku, Finland. Link{\"o}ping University
  Electronic Press.

\bibitem[{Emerson(2020)}]{emerson-2020-goals}
Guy Emerson. 2020.
\newblock \href {https://doi.org/10.18653/v1/2020.acl-main.663} {What are the
  goals of distributional semantics?}
\newblock In \emph{Proceedings of the 58th Annual Meeting of the Association
  for Computational Linguistics}, pages 7436--7453, Online. Association for
  Computational Linguistics.

\bibitem[{Firth(1957)}]{firth_synopsis_1957}
John~R Firth. 1957.
\newblock A synopsis of linguistic theory, 1930-1955.
\newblock \emph{Studies in linguistic analysis}.

\bibitem[{Flickinger(2000)}]{flickinger-2000-building}
Dan Flickinger. 2000.
\newblock On building a more effcient grammar by exploiting types.
\newblock \emph{Natural Language Engineering}, 6(1):15--28.

\bibitem[{Geiger et~al.(2021)Geiger, Lu, Icard, and
  Potts}]{Geiger:Lu-etal:2021-abstractions}
Atticus Geiger, Hanson Lu, Thomas Icard, and Christopher Potts. 2021.
\newblock \href {https://arxiv.org/abs/2106.02997} {Causal abstractions of
  neural networks}.
\newblock In \emph{Proc. of NeurIPS}.

\bibitem[{Glava{\v{s}} and Vuli{\'c}(2021)}]{glavas-vulic-2021-supervised}
Goran Glava{\v{s}} and Ivan Vuli{\'c}. 2021.
\newblock \href {https://doi.org/10.18653/v1/2021.eacl-main.270} {Is supervised
  syntactic parsing beneficial for language understanding tasks? an empirical
  investigation}.
\newblock In \emph{Proceedings of the 16th Conference of the European Chapter
  of the Association for Computational Linguistics: Main Volume}, pages
  3090--3104, Online. Association for Computational Linguistics.

\bibitem[{Haji{\v{c}} et~al.(2012)Haji{\v{c}}, Haji{\v{c}}ov{\'a},
  Panevov{\'a}, Sgall, Bojar, Cinkov{\'a}, Fu{\v{c}}{\'\i}kov{\'a},
  Mikulov{\'a}, Pajas, Popelka, Semeck{\'y}, {\v{S}}indlerov{\'a},
  {\v{S}}t{\v{e}}p{\'a}nek, Toman, Ure{\v{s}}ov{\'a}, and
  {\v{Z}}abokrtsk{\'y}}]{hajic-etal-2012-announcing}
Jan Haji{\v{c}}, Eva Haji{\v{c}}ov{\'a}, Jarmila Panevov{\'a}, Petr Sgall,
  Ond{\v{r}}ej Bojar, Silvie Cinkov{\'a}, Eva Fu{\v{c}}{\'\i}kov{\'a}, Marie
  Mikulov{\'a}, Petr Pajas, Jan Popelka, Ji{\v{r}}{\'\i} Semeck{\'y}, Jana
  {\v{S}}indlerov{\'a}, Jan {\v{S}}t{\v{e}}p{\'a}nek, Josef Toman, Zde{\v{n}}ka
  Ure{\v{s}}ov{\'a}, and Zden{\v{e}}k {\v{Z}}abokrtsk{\'y}. 2012.
\newblock \href
  {http://www.lrec-conf.org/proceedings/lrec2012/pdf/510_Paper.pdf} {Announcing
  {P}rague {C}zech-{E}nglish {D}ependency {T}reebank 2.0}.
\newblock In \emph{Proceedings of the Eighth International Conference on
  Language Resources and Evaluation ({LREC}'12)}, pages 3153--3160, Istanbul,
  Turkey. European Language Resources Association (ELRA).

\bibitem[{Hale et~al.(2018)Hale, Dyer, Kuncoro, and
  Brennan}]{hale-etal-2018-finding}
John Hale, Chris Dyer, Adhiguna Kuncoro, and Jonathan Brennan. 2018.
\newblock \href {https://doi.org/10.18653/v1/P18-1254} {Finding syntax in human
  encephalography with beam search}.
\newblock In \emph{Proceedings of the 56th Annual Meeting of the Association
  for Computational Linguistics (Volume 1: Long Papers)}, pages 2727--2736,
  Melbourne, Australia. Association for Computational Linguistics.

\bibitem[{Hao et~al.(2020)Hao, Mendelsohn, Sterneck, Martinez, and
  Frank}]{hao2020probabilistic}
Yiding Hao, Simon Mendelsohn, Rachel Sterneck, Randi Martinez, and Robert
  Frank. 2020.
\newblock {Probabilistic Predictions of People Perusing: Evaluating Metrics of
  Language Model Performance for Psycholinguistic Modeling}.
\newblock In \emph{Proceedings of the EMNLP Workshop on Cognitive Modeling and
  Computational Linguistics}.

\bibitem[{Harris(1954)}]{harris_distributional_1954}
Zellig~S Harris. 1954.
\newblock Distributional structure.
\newblock \emph{Word}, 10(2-3):146--162.

\bibitem[{Hassan et~al.(2009)Hassan, Sima{'}an, and
  Way}]{hassan-etal-2009-lexicalized}
Hany Hassan, Khalil Sima{'}an, and Andy Way. 2009.
\newblock \href {https://aclanthology.org/R09-1025} {Lexicalized
  semi-incremental dependency parsing}.
\newblock In \emph{Proceedings of the International Conference {RANLP}-2009},
  pages 128--134, Borovets, Bulgaria. Association for Computational
  Linguistics.

\bibitem[{Hewitt and Liang(2019)}]{hewitt-liang-2019-designing}
John Hewitt and Percy Liang. 2019.
\newblock \href {https://doi.org/10.18653/v1/D19-1275} {Designing and
  interpreting probes with control tasks}.
\newblock In \emph{Proceedings of the 2019 Conference on Empirical Methods in
  Natural Language Processing and the 9th International Joint Conference on
  Natural Language Processing (EMNLP-IJCNLP)}, pages 2733--2743, Hong Kong,
  China. Association for Computational Linguistics.

\bibitem[{Hu et~al.(2020)Hu, Gauthier, Qian, Wilcox, and
  Levy}]{hu-etal-2020-systematic}
Jennifer Hu, Jon Gauthier, Peng Qian, Ethan Wilcox, and Roger Levy. 2020.
\newblock \href {https://doi.org/10.18653/v1/2020.acl-main.158} {A systematic
  assessment of syntactic generalization in neural language models}.
\newblock In \emph{Proceedings of the 58th Annual Meeting of the Association
  for Computational Linguistics}, pages 1725--1744, Online. Association for
  Computational Linguistics.

\bibitem[{Jacovi et~al.(2021)Jacovi, Swayamdipta, Ravfogel, Elazar, Choi, and
  Goldberg}]{jacovi-etal-2021-contrastive}
Alon Jacovi, Swabha Swayamdipta, Shauli Ravfogel, Yanai Elazar, Yejin Choi, and
  Yoav Goldberg. 2021.
\newblock \href {https://doi.org/10.18653/v1/2021.emnlp-main.120} {Contrastive
  explanations for model interpretability}.
\newblock In \emph{Proceedings of the 2021 Conference on Empirical Methods in
  Natural Language Processing}, pages 1597--1611, Online and Punta Cana,
  Dominican Republic. Association for Computational Linguistics.

\bibitem[{Jawahar et~al.(2019)Jawahar, Sagot, and
  Seddah}]{jawahar-etal-2019-bert}
Ganesh Jawahar, Beno{\^\i}t Sagot, and Djam{\'e} Seddah. 2019.
\newblock \href {https://doi.org/10.18653/v1/P19-1356} {What does {BERT} learn
  about the structure of language?}
\newblock In \emph{Proceedings of the 57th Annual Meeting of the Association
  for Computational Linguistics}, pages 3651--3657, Florence, Italy.
  Association for Computational Linguistics.

\bibitem[{Kim et~al.(2019)Kim, Patel, Poliak, Xia, Wang, McCoy, Tenney, Ross,
  Linzen, Van~Durme, Bowman, and Pavlick}]{kim-etal-2019-probing}
Najoung Kim, Roma Patel, Adam Poliak, Patrick Xia, Alex Wang, Tom McCoy, Ian
  Tenney, Alexis Ross, Tal Linzen, Benjamin Van~Durme, Samuel~R. Bowman, and
  Ellie Pavlick. 2019.
\newblock \href {https://doi.org/10.18653/v1/S19-1026} {Probing what different
  {NLP} tasks teach machines about function word comprehension}.
\newblock In \emph{Proceedings of the Eighth Joint Conference on Lexical and
  Computational Semantics (*{SEM} 2019)}, pages 235--249, Minneapolis,
  Minnesota. Association for Computational Linguistics.

\bibitem[{Kingma and Welling(2013)}]{kingma2013auto}
Diederik~P Kingma and Max Welling. 2013.
\newblock Auto-encoding variational bayes.
\newblock \emph{arXiv preprint arXiv:1312.6114}.

\bibitem[{Kiperwasser and Goldberg(2016)}]{kiperwasser-goldberg-2016-simple}
Eliyahu Kiperwasser and Yoav Goldberg. 2016.
\newblock \href {https://doi.org/10.1162/tacl_a_00101} {Simple and accurate
  dependency parsing using bidirectional {LSTM} feature representations}.
\newblock \emph{Transactions of the Association for Computational Linguistics},
  4:313--327.

\bibitem[{Kitaev et~al.(2022)Kitaev, Lu, and Klein}]{kitaev-etal-2022-learned}
Nikita Kitaev, Thomas Lu, and Dan Klein. 2022.
\newblock \href {https://doi.org/10.18653/v1/2022.acl-long.220} {Learned
  incremental representations for parsing}.
\newblock In \emph{Proceedings of the 60th Annual Meeting of the Association
  for Computational Linguistics (Volume 1: Long Papers)}, pages 3086--3095,
  Dublin, Ireland. Association for Computational Linguistics.

\bibitem[{Lenci(2023)}]{lenci2023understanding}
Alessandro Lenci. 2023.
\newblock \href {https://arxiv.org/abs/2303.04229} {Understanding natural
  language understanding systems. a critical analysis}.
\newblock Preprint arXiv:2303.04229.

\bibitem[{Li and Srikumar(2019)}]{li-srikumar-2019-augmenting}
Tao Li and Vivek Srikumar. 2019.
\newblock \href {https://doi.org/10.18653/v1/P19-1028} {Augmenting neural
  networks with first-order logic}.
\newblock In \emph{Proceedings of the 57th Annual Meeting of the Association
  for Computational Linguistics}, pages 292--302, Florence, Italy. Association
  for Computational Linguistics.

\bibitem[{Li and Rush(2020)}]{li-rush-2020-posterior}
Xiang~Lisa Li and Alexander Rush. 2020.
\newblock \href {https://doi.org/10.18653/v1/2020.acl-main.243} {Posterior
  control of blackbox generation}.
\newblock In \emph{Proceedings of the 58th Annual Meeting of the Association
  for Computational Linguistics}, pages 2731--2743, Online. Association for
  Computational Linguistics.

\bibitem[{Linzen et~al.(2016)Linzen, Dupoux, and
  Goldberg}]{linzen-etal-2016-assessing}
Tal Linzen, Emmanuel Dupoux, and Yoav Goldberg. 2016.
\newblock \href {https://doi.org/10.1162/tacl_a_00115} {Assessing the ability
  of {LSTM}s to learn syntax-sensitive dependencies}.
\newblock \emph{Transactions of the Association for Computational Linguistics},
  4:521--535.

\bibitem[{Liu et~al.(2019)Liu, Gardner, Belinkov, Peters, and
  Smith}]{liu-etal-2019-linguistic}
Nelson~F. Liu, Matt Gardner, Yonatan Belinkov, Matthew~E. Peters, and Noah~A.
  Smith. 2019.
\newblock \href {https://doi.org/10.18653/v1/N19-1112} {Linguistic knowledge
  and transferability of contextual representations}.
\newblock In \emph{Proceedings of the 2019 Conference of the North {A}merican
  Chapter of the Association for Computational Linguistics: Human Language
  Technologies, Volume 1 (Long and Short Papers)}, pages 1073--1094,
  Minneapolis, Minnesota. Association for Computational Linguistics.

\bibitem[{Merrill et~al.(2021)Merrill, Goldberg, Schwartz, and
  Smith}]{merrill-etal-2021-provable}
William Merrill, Yoav Goldberg, Roy Schwartz, and Noah~A. Smith. 2021.
\newblock \href {https://doi.org/10.1162/tacl_a_00412} {Provable limitations of
  acquiring meaning from ungrounded form: What will future language models
  understand?}
\newblock \emph{Transactions of the Association for Computational Linguistics},
  9:1047--1060.

\bibitem[{Oepen et~al.(2020)Oepen, Abend, Abzianidze, Bos, Hajic, Hershcovich,
  Li, O{'}Gorman, Xue, and Zeman}]{oepen-etal-2020-mrp}
Stephan Oepen, Omri Abend, Lasha Abzianidze, Johan Bos, Jan Hajic, Daniel
  Hershcovich, Bin Li, Tim O{'}Gorman, Nianwen Xue, and Daniel Zeman. 2020.
\newblock \href {https://doi.org/10.18653/v1/2020.conll-shared.1} {{MRP} 2020:
  The second shared task on cross-framework and cross-lingual meaning
  representation parsing}.
\newblock In \emph{Proceedings of the CoNLL 2020 Shared Task: Cross-Framework
  Meaning Representation Parsing}, pages 1--22, Online. Association for
  Computational Linguistics.

\bibitem[{Oepen et~al.(2019)Oepen, Abend, Hajic, Hershcovich, Kuhlmann,
  O{'}Gorman, Xue, Chun, Straka, and Uresova}]{oepen-etal-2019-mrp}
Stephan Oepen, Omri Abend, Jan Hajic, Daniel Hershcovich, Marco Kuhlmann, Tim
  O{'}Gorman, Nianwen Xue, Jayeol Chun, Milan Straka, and Zdenka Uresova. 2019.
\newblock \href {https://doi.org/10.18653/v1/K19-2001} {{MRP} 2019:
  Cross-framework meaning representation parsing}.
\newblock In \emph{Proceedings of the Shared Task on Cross-Framework Meaning
  Representation Parsing at the 2019 Conference on Natural Language Learning},
  pages 1--27, Hong Kong. Association for Computational Linguistics.

\bibitem[{Oepen and L{\o}nning(2006)}]{oepen-lonning-2006-discriminant}
Stephan Oepen and Jan~Tore L{\o}nning. 2006.
\newblock \href {http://www.lrec-conf.org/proceedings/lrec2006/pdf/364_pdf.pdf}
  {Discriminant-based {MRS} banking}.
\newblock In \emph{Proceedings of the Fifth International Conference on
  Language Resources and Evaluation ({LREC}{'}06)}, Genoa, Italy. European
  Language Resources Association (ELRA).

\bibitem[{Opitz and Frank(2022)}]{opitz-frank-2022-sbert}
Juri Opitz and Anette Frank. 2022.
\newblock \href {https://aclanthology.org/2022.aacl-main.48} {{SBERT} studies
  meaning representations: Decomposing sentence embeddings into explainable
  semantic features}.
\newblock In \emph{Proceedings of the 2nd Conference of the Asia-Pacific
  Chapter of the Association for Computational Linguistics and the 12th
  International Joint Conference on Natural Language Processing (Volume 1: Long
  Papers)}, pages 625--638, Online only. Association for Computational
  Linguistics.

\bibitem[{Polino et~al.(2018)Polino, Pascanu, and Alistarh}]{polino2018model}
Antonio Polino, Razvan Pascanu, and Dan Alistarh. 2018.
\newblock \href {https://openreview.net/forum?id=S1XolQbRW} {Model compression
  via distillation and quantization}.
\newblock In \emph{International Conference on Learning Representations}.

\bibitem[{Prange et~al.(2022)Prange, Schneider, and
  Kong}]{prange-etal-2022-linguistic}
Jakob Prange, Nathan Schneider, and Lingpeng Kong. 2022.
\newblock \href {https://doi.org/10.18653/v1/2022.naacl-main.325} {Linguistic
  frameworks go toe-to-toe at neuro-symbolic language modeling}.
\newblock In \emph{Proceedings of the 2022 Conference of the North American
  Chapter of the Association for Computational Linguistics: Human Language
  Technologies}, pages 4375--4391, Seattle, United States. Association for
  Computational Linguistics.

\bibitem[{Prange et~al.(2021)Prange, Schneider, and
  Srikumar}]{prange-etal-2021-supertagging}
Jakob Prange, Nathan Schneider, and Vivek Srikumar. 2021.
\newblock \href {https://doi.org/10.1162/tacl_a_00364} {Supertagging the long
  tail with tree-structured decoding of complex categories}.
\newblock \emph{Transactions of the Association for Computational Linguistics},
  9:243--260.

\bibitem[{Qian et~al.(2021)Qian, Naseem, Levy, and
  Fernandez~Astudillo}]{qian-etal-2021-structural}
Peng Qian, Tahira Naseem, Roger Levy, and Ram{\'o}n Fernandez~Astudillo. 2021.
\newblock \href {https://doi.org/10.18653/v1/2021.acl-long.289} {Structural
  guidance for transformer language models}.
\newblock In \emph{Proceedings of the 59th Annual Meeting of the Association
  for Computational Linguistics and the 11th International Joint Conference on
  Natural Language Processing (Volume 1: Long Papers)}, pages 3735--3745,
  Online. Association for Computational Linguistics.

\bibitem[{Radford et~al.(2019)Radford, Wu, Child, Luan, Amodei, Sutskever
  et~al.}]{radford2019language}
Alec Radford, Jeffrey Wu, Rewon Child, David Luan, Dario Amodei, Ilya
  Sutskever, et~al. 2019.
\newblock \href
  {https://cdn.openai.com/better-language-models/language_models_are_unsupervised_multitask_learners.pdf}
  {Language models are unsupervised multitask learners}.
\newblock OpenAI blog.

\bibitem[{Sanh et~al.(2019)Sanh, Debut, Chaumond, and
  Wolf}]{sanh2019distilbert}
Victor Sanh, Lysandre Debut, Julien Chaumond, and Thomas Wolf. 2019.
\newblock \href {https://arxiv.org/abs/1910.01108} {Distilbert, a distilled
  version of bert: smaller, faster, cheaper and lighter}.
\newblock Preprint arXiv:1910.01108.

\bibitem[{Sartran et~al.(2022)Sartran, Barrett, Kuncoro, Stanojević, Blunsom,
  and Dyer}]{sartran-etal-2022-transformer}
Laurent Sartran, Samuel Barrett, Adhiguna Kuncoro, Miloš Stanojević, Phil
  Blunsom, and Chris Dyer. 2022.
\newblock \href {https://doi.org/10.1162/tacl_a_00526} {{Transformer Grammars:
  Augmenting Transformer Language Models with Syntactic Inductive Biases at
  Scale}}.
\newblock \emph{Transactions of the Association for Computational Linguistics},
  10:1423--1439.

\bibitem[{Schlichtkrull et~al.(2018)Schlichtkrull, Kipf, Bloem, van den Berg,
  Titov, and Welling}]{schlichtkrull-etal-2018-rgcn}
Michael Schlichtkrull, Thomas~N. Kipf, Peter Bloem, Rianne van den Berg, Ivan
  Titov, and Max Welling. 2018.
\newblock \href
  {https://link.springer.com/chapter/10.1007\%2F978-3-319-93417-4_38} {Modeling
  relational data with graph convolutional networks}.
\newblock In \emph{The Semantic Web}, pages 593--607, Cham. Springer
  International Publishing.

\bibitem[{Sgall et~al.(1986)Sgall, Hajicov{\'a}, and
  Panevov{\'a}}]{sgall-etal-1986-meaning}
Petr Sgall, Eva Hajicov{\'a}, and Jarmila Panevov{\'a}. 1986.
\newblock The meaning of the sentence and its semantic and pragmatic aspects.
  academia.

\bibitem[{Stanojevi{\'c} et~al.(2021)Stanojevi{\'c}, Bhattasali, Dunagan,
  Campanelli, Steedman, Brennan, and Hale}]{stanojevic-etal-2021-modeling}
Milo{\v{s}} Stanojevi{\'c}, Shohini Bhattasali, Donald Dunagan, Luca
  Campanelli, Mark Steedman, Jonathan Brennan, and John Hale. 2021.
\newblock \href {https://doi.org/10.18653/v1/2021.cmcl-1.3} {Modeling
  incremental language comprehension in the brain with {C}ombinatory
  {C}ategorial {G}rammar}.
\newblock In \emph{Proceedings of the Workshop on Cognitive Modeling and
  Computational Linguistics}, pages 23--38, Online. Association for
  Computational Linguistics.

\bibitem[{Stanojevi{\'c} and Steedman(2019)}]{stanojevic-steedman-2019-ccg}
Milo{\v{s}} Stanojevi{\'c} and Mark Steedman. 2019.
\newblock \href {https://doi.org/10.18653/v1/N19-1020} {{CCG} parsing algorithm
  with incremental tree rotation}.
\newblock In \emph{Proceedings of the 2019 Conference of the North {A}merican
  Chapter of the Association for Computational Linguistics: Human Language
  Technologies, Volume 1 (Long and Short Papers)}, pages 228--239, Minneapolis,
  Minnesota. Association for Computational Linguistics.

\bibitem[{Stanojevi{\'c} and Steedman(2020)}]{stanojevic-steedman-2020-max}
Milo{\v{s}} Stanojevi{\'c} and Mark Steedman. 2020.
\newblock \href {https://doi.org/10.18653/v1/2020.acl-main.378} {Max-margin
  incremental {CCG} parsing}.
\newblock In \emph{Proceedings of the 58th Annual Meeting of the Association
  for Computational Linguistics}, pages 4111--4122, Online. Association for
  Computational Linguistics.

\bibitem[{Stern et~al.(2017)Stern, Fried, and
  Klein}]{stern-etal-2017-effective}
Mitchell Stern, Daniel Fried, and Dan Klein. 2017.
\newblock \href {https://doi.org/10.18653/v1/D17-1178} {Effective inference for
  generative neural parsing}.
\newblock In \emph{Proceedings of the 2017 Conference on Empirical Methods in
  Natural Language Processing}, pages 1695--1700, Copenhagen, Denmark.
  Association for Computational Linguistics.

\bibitem[{Tenney et~al.(2019{\natexlab{a}})Tenney, Das, and
  Pavlick}]{tenney-etal-2019-bert}
Ian Tenney, Dipanjan Das, and Ellie Pavlick. 2019{\natexlab{a}}.
\newblock \href {https://doi.org/10.18653/v1/P19-1452} {{BERT} rediscovers the
  classical {NLP} pipeline}.
\newblock In \emph{Proceedings of the 57th Annual Meeting of the Association
  for Computational Linguistics}, pages 4593--4601, Florence, Italy.
  Association for Computational Linguistics.

\bibitem[{Tenney et~al.(2019{\natexlab{b}})Tenney, Xia, Chen, Wang, Poliak,
  McCoy, Kim, Durme, Bowman, Das, and Pavlick}]{tenney2018what}
Ian Tenney, Patrick Xia, Berlin Chen, Alex Wang, Adam Poliak, R~Thomas McCoy,
  Najoung Kim, Benjamin~Van Durme, Sam Bowman, Dipanjan Das, and Ellie Pavlick.
  2019{\natexlab{b}}.
\newblock \href {https://openreview.net/forum?id=SJzSgnRcKX} {What do you learn
  from context? probing for sentence structure in contextualized word
  representations}.
\newblock In \emph{Proc. of ICLR}.

\bibitem[{Trott et~al.(2020)Trott, Torrent, Chang, and
  Schneider}]{trott-etal-2020-construing}
Sean Trott, Tiago~Timponi Torrent, Nancy Chang, and Nathan Schneider. 2020.
\newblock \href {https://doi.org/10.18653/v1/2020.acl-main.462}
  {({R}e)construing meaning in {NLP}}.
\newblock In \emph{Proceedings of the 58th Annual Meeting of the Association
  for Computational Linguistics}, pages 5170--5184, Online. Association for
  Computational Linguistics.

\bibitem[{Warstadt et~al.(2020)Warstadt, Parrish, Liu, Mohananey, Peng, Wang,
  and Bowman}]{warstadt-etal-2020-blimp-benchmark}
Alex Warstadt, Alicia Parrish, Haokun Liu, Anhad Mohananey, Wei Peng, Sheng-Fu
  Wang, and Samuel~R. Bowman. 2020.
\newblock \href {https://doi.org/10.1162/tacl_a_00321} {{BL}i{MP}: The
  benchmark of linguistic minimal pairs for {E}nglish}.
\newblock \emph{Transactions of the Association for Computational Linguistics},
  8:377--392.

\bibitem[{Wei{\ss}enhorn et~al.(2022)Wei{\ss}enhorn, Donatelli, and
  Koller}]{weissenhorn-etal-2022-compositional}
Pia Wei{\ss}enhorn, Lucia Donatelli, and Alexander Koller. 2022.
\newblock \href {https://doi.org/10.18653/v1/2022.starsem-1.4} {Compositional
  generalization with a broad-coverage semantic parser}.
\newblock In \emph{Proceedings of the 11th Joint Conference on Lexical and
  Computational Semantics}, pages 44--54, Seattle, Washington. Association for
  Computational Linguistics.

\bibitem[{Wu et~al.(2020)Wu, Chen, Kao, and Liu}]{wu-etal-2020-perturbed}
Zhiyong Wu, Yun Chen, Ben Kao, and Qun Liu. 2020.
\newblock \href {https://doi.org/10.18653/v1/2020.acl-main.383} {Perturbed
  masking: Parameter-free probing for analyzing and interpreting {BERT}}.
\newblock In \emph{Proceedings of the 58th Annual Meeting of the Association
  for Computational Linguistics}, pages 4166--4176, Online. Association for
  Computational Linguistics.

\bibitem[{Yin and Neubig(2022)}]{yin-neubig-2022-interpreting}
Kayo Yin and Graham Neubig. 2022.
\newblock \href {https://aclanthology.org/2022.emnlp-main.14} {Interpreting
  language models with contrastive explanations}.
\newblock In \emph{Proceedings of the 2022 Conference on Empirical Methods in
  Natural Language Processing}, pages 184--198, Abu Dhabi, United Arab
  Emirates. Association for Computational Linguistics.

\end{thebibliography}
\bibliographystyle{acl_natbib}

\end{document}